\documentclass[times, review, 10pt]{elsarticle}
\usepackage[letterpaper, top=4.3cm, bottom=4.3cm, left=4.8cm, right=4.8cm]{geometry}

\usepackage{amsmath,amssymb,amsfonts}
\usepackage{algorithmic, algorithm}
\usepackage{graphicx}
\usepackage{hyperref}
\usepackage{booktabs}
\usepackage{multirow}
\usepackage{gensymb}
\usepackage{tabularx} 
\usepackage{tablefootnote}
\hypersetup{hidelinks=true}
\usepackage[dvipsnames]{xcolor}
\usepackage{textcomp}
\usepackage{pdflscape}
\usepackage{threeparttable}
\usepackage{color,soul}
\usepackage{lineno} 

\journal{Pattern Recognition}

\begin{document}

\begin{frontmatter}

\title{Volumetric Imaging-guided Multimodal In-bed 3D Pose and Shape Estimation}

\author[sydney_cs]{Mingxiao Tu}
\author[sydney_cs]{Hoijoon Jung}
\author[westmead]{Alireza Moghadam}
\author[westmead]{Jineel Raythatha}
\author[westmead]{Lachlan Allan}
\author[westmead]{Jeremy Hsu}
\author[sydney_bme]{Andre Kyme\corref{cor1}}
\author[sydney_cs]{Jinman Kim\corref{cor1}}

\cortext[cor1]{Corresponding authors. E-mail addresses: andre.kyme@sydney.edu.au (A. Kyme), jinman.kim@sydney.edu.au (J. Kim).}

\address[sydney_cs]{School of Computer Science, The University of Sydney, Sydney, NSW 2006, Australia}
\address[sydney_bme]{School of Biomedical Engineering, The University of Sydney, Sydney, NSW 2006, Australia}
\address[westmead]{Trauma Service, Westmead Hospital, Australia}

\begin{abstract}
Accurate in-bed 3D patient pose and shape estimation (PSE) is crucial for clinical applications in perioperative care, including high-precision Augmented Reality (AR)-guided surgical navigation and proactive pressure injury prevention. However, reliable in-bed PSE remains challenging due to the combination of severe occlusion from drapes and the unavailability of patient-specific anatomical geometry. Conventional methods relying on depth, RGB, or pressure maps are limited by line-of-sight when the body is obscured, preventing the reconstruction of shape and pose beneath the occluding drapery. Attempts to resolve this through synthetic data training often suffer from the `reality gap' in modelling complex drape deformations, while recent domain adaptation techniques rely on statistical population priors that cannot capture patient-specific anatomy. Consequently, these approaches fail to reliably model patients with atypical anatomies under occlusion, compromising reconstruction reliability and hindering adoption in high-precision clinical systems.

To overcome these limitations, we propose Volumetric Imaging-guided Multimodal In-bed 3D Pose and Shape Estimation (VIM-PSE), a novel multimodal framework that integrates depth image data with a high-fidelity, patient-specific anatomical prior derived from volumetric medical imaging (e.g., CT or MRI). By leveraging routinely acquired medical imaging in the perioperative workflow, our method resolves the pose and shape ambiguity for individual patients under occlusion. VIM-PSE proposes a robust, lightweight Cross-modal Residual Fusion (CRF) module to fuse the two input modalities. We validated VIM-PSE on a large-scale MRI-based simulation dataset (N=300) to confirm statistical robustness across a diverse population, and on CT-based real-world phantom (N=1) and in-vivo volunteer datasets (N=6) to demonstrate clinical feasibility. VIM-PSE consistently surpassed state-of-the-art baselines across shape and pose estimation metrics, achieving a $>$49\% average improvement in shape estimation under occlusion compared with baselines. Crucially, our method demonstrated a torso vertex-to-vertex error of 0.26 cm, validating its capability to meet the stringent high-precision requirements for safety-critical clinical guidance systems. Code and the phantom dataset will be released upon publication to support transparent reproduction and further research.
\end{abstract}

\begin{keyword}
3D Pose and Shape Estimation \sep Volumetric Medical Imaging \sep Multimodal Fusion
\end{keyword}

\end{frontmatter}


\section{Introduction}
\label{sec:introduction}
In the hospital environment, perioperative care encompasses clinical support provided before (preoperative), during (intraoperative), and after (postoperative) surgery \citep{meara2015global}. A notable portion of patients remain confined to bed during these stages. Approximately 15\% of patients are bedridden preoperatively, and nearly 100\% spend the entire surgical procedure on the operating table (typically lasting 2-6 hours) \citep{safety2009guidelines}. Studies report that 30–40\% of patients remain in bed postoperatively for recovery periods ranging from a few days to several weeks, depending on the type of operation \citep{collins1999risk}.

The inherent constraints of the perioperative environment, where patients are immobilized and often obscured by bedding or surgical drapes, introduce critical challenges regarding patient safety and surgical precision. First, prolonged immobility contributes to adverse outcomes such as pressure ulcers, affecting approximately 2.5 million patients annually in the United States \citep{berlowitz2011preventing}. Second, precise patient positioning is essential for surgical success; for example, during surgery, accurate alignment is required to overlay patient-specific medical images onto the body (`X-ray'-like visualization) to support surgical navigation and decision-making \citep{jung2024ribmr}. Third, postoperatively, identifying deviations from prescribed positions is necessary to promote optimal wound healing or respiratory mechanics \citep{casas2019patient,chen2018patient}. Addressing these, as well as other diverse clinical requirements, necessitates accurate three-dimensional (3D) patient pose and body shape estimation (PSE) capable of functioning in this constrained environment.

\begin{figure}
\centering
\includegraphics[width=0.8\textwidth]{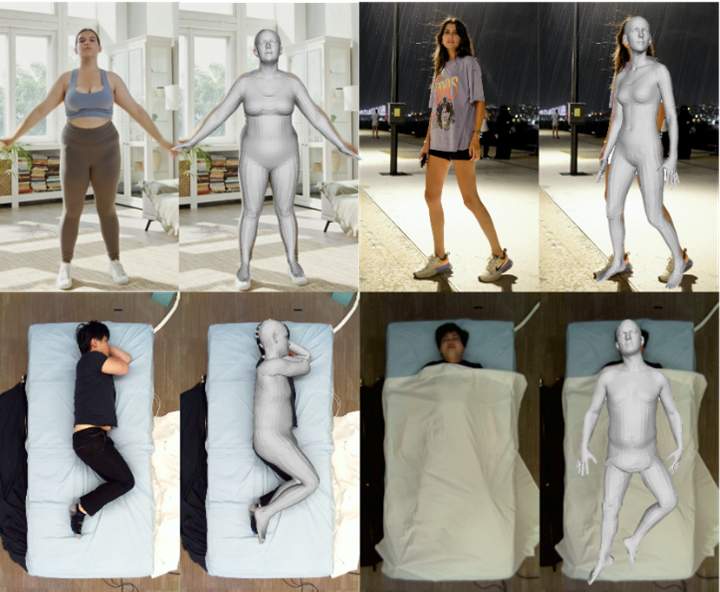}
\caption{3D PSE in general multimedia field (first row) \citep{meshcapade_demo} compared to the perioperative setting (second row) \citep{liu2022simultaneously}, where the latter faces challenges from occlusions of body parts such as the surgical drapes, and non-standard poses. The predicted SMPL mesh, a 3D human mesh model, is overlaid on the original image as a grey mesh.} 
\label{fig1}
\end{figure}

The foundation for modern PSE solutions is the parametric Skinned Multi-Person Linear (SMPL) model \citep{SMPL:2015}. This model represents the human body as a 3D mesh by decoupling shape and pose into low-dimensional parameter spaces as shown in Fig~\ref{fig1}. Existing in-bed patient PSE methods typically rely on end-to-end deep learning networks that estimate these SMPL parameters from one or more input modalities to generate a 3D patient mesh model  \citep{yin2022multimodal,clever20183d,clever2022bodypressure,tandon2024bodymap,clever2020bodies}. \citet{clever2020bodies} presented a deep learning model that infers SMPL parameters based on pressure maps (PM) and gender information. \citet{yin2022multimodal} incorporated multimodal data (RGB-D, IR, and PM) to improve in-bed PSE performance. \citet{zheng2022self} introduced a self-supervised deep learning network that uses 2D keypoint estimation from RGB-D images. More recently, \citet{tandon2024bodymap} proposed BodyMAP that jointly predicts a SMPL model and a 3D pressure map from depth and 2D pressure images.

Despite these advancements, the robustness and accuracy of current PSE methods are hindered by occlusions from bedding or medical drapes, and the variability of complex in-bed patient poses. To address the challenge of occlusion without relying on large-scale labelled data of covered patients, recent research has explored domain adaptation techniques. For instance, \citet{bigalke2023anatomy} used anatomy-guided domain adaptation to bridge the gap between uncovered training data and the covered test domain. However, while effective for domain alignment, these approaches inherently rely on generic regularization constraints (such as limb symmetry or standard proportions) to resolve the ambiguity of occluded regions. Consequently, when visual cues are missing, these methods tend to bias predictions toward a healthy statistical norm, failing to reliably capture patient-specific pathologies or atypical anatomies that naturally deviate from these standard constraints.

To address these limitations, we introduce Volumetric Imaging-guided Multimodal In-bed 3D Pose and Shape Estimation (VIM-PSE). By leveraging volumetric scans, such as CT and MRI, routinely acquired for perioperative assessment \citep{merlo2017does}, intraoperative navigation \citep{uhl2009intraoperative}, or postoperative evaluation \citep{heary2004thoracic}, VIM-PSE establishes a reliable `reference view' of the patient's internal geometry. Our method extracts detailed surface features from volumetric data through a shape estimation module, and combines this with a human pose estimation module, which derives pose from depth maps. To further improve cross-modality fusion, we introduce a lightweight trainable Cross-modal Residual Fusion Module (CRF) that allows features from volumetric data and depth modalities to guide and refine each other through confidence-weighted residual corrections. This multimodal approach effectively compensates for the loss of surface information caused by surgical drapes, enabling the reconstruction of a complete and anatomically precise 3D patient model even when external visual cues are obscured. We evaluate our framework's cross-modality generalizability on the large-scale MRI-based public HIT dataset \citep{keller2024hit}, and on proprietary CT-based phantom and volunteer datasets to demonstrate clinical feasibility.

\begin{figure}
\centering
\includegraphics[width=0.85\textwidth]{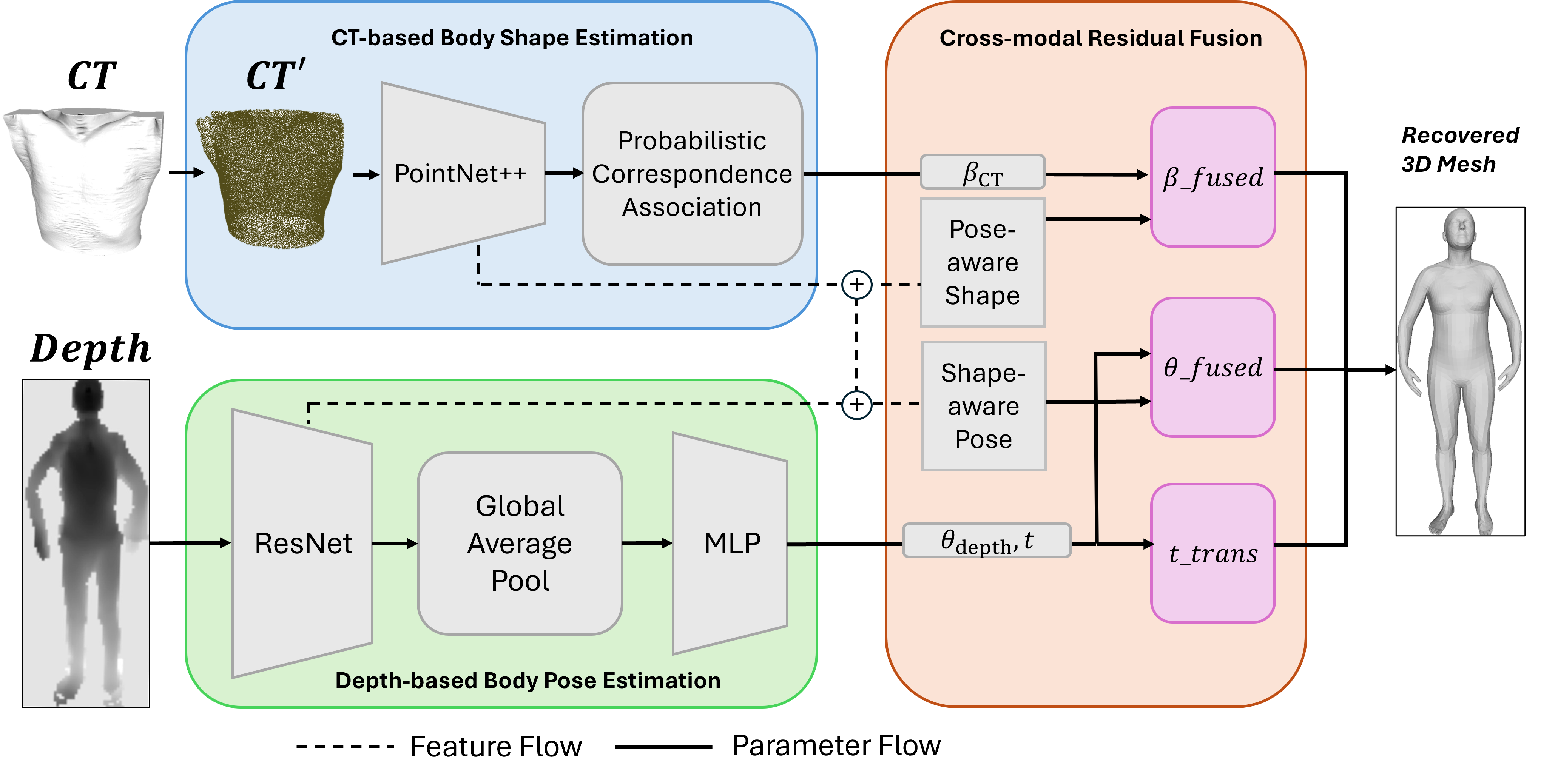}
\caption{Our proposed VIM-PSE framework, which takes volumetric data and depth image as inputs to infer a SMPL model. Module 1 uses PointNet++ to extract features from the point cloud derived from CT or MRI surface, and shape parameters are then inferred through Probabilistic Correspondence Association (PCA). These shape parameters are retained for the final SMPL model. Module 2 uses depth map to infer pose parameters through ResNet and MLP. The Cross-modal Confidence-Guided Module (CRF) enables mutual refinement between shape and pose features through residual corrections. This combines outputs from both modalities to infer the final SMPL model.} 
\label{fig2}
\end{figure}

Our contributions are threefold:
\begin{enumerate}
    \item We present the first, to our knowledge, \emph{in-bed} PSE pipeline that fuses a volumetric torso shape prior with depth-based pose estimation. Crucially, our framework is explicitly validated on clinically relevant in-bed postures (e.g., supine, lateral), enabling robust reconstruction of complex body configurations even under bedding occlusion.
    \item We provide a clinically grounded evaluation achieving a torso-specific vertex-to-vertex (V2V) error of 0.26\,cm, which falls within the accuracy requirements for common augmented-reality (AR) surgical procedures such as pedicle screw placement \citep{ElmiTerander2016_AR_SPINE, Spirig2021_AR_Pedicle}, validated on phantom and in-vivo volunteer data.
    \item We have collected and will release a in-bed phantom dataset containing paired 3D surface scans, depth images, CT volumes, metadata and registered SMPL ground truth, establishing a high-fidelity benchmark to facilitate future research in multimodal in-bed PSE and its applications.
\end{enumerate}

\section{Related Work}
\label{sec:related_work}
\subsection{In-bed Human Pose and Shape Estimation}
Human Pose Estimation (HPE) is a fundamental computer vision task, typically categorized into 2D keypoint localization \citep{sun2019deep, xiao2018simple} and 3D Pose and Shape Estimation (PSE). The latter aims to reconstruct the full body surface, predominantly by regressing parameters of the Skinned Multi-Person Linear (SMPL) model \citep{SMPL:2015}, which effectively decouples body geometry into low-dimensional shape ($\beta$) and pose ($\theta$) components.

While general HPE focuses on upright subjects in open scenes, the in-bed patient setting presents unique challenges, including difficult patient positions (typically supine or lateral positioning) and environmental factors such as surgical drapes. Unimodal approaches have attempted to solve this using specific sensors. Thermal imaging \citep{liu2019seeing} and Pressure Maps (PM) \citep{casas2019patient, davoodnia2021low} offer privacy-preserving modalities, while depth-based methods \citep{bigalke2023anatomy} utilise 3D point clouds to capture geometry. However, unimodal methods face a fundamental bottleneck: surface-based sensors lack volumetric data, and 3D estimation from a single view is an ill-posed problem, particularly when the patient is occluded, necessitating strong priors to resolve the ambiguity.

\subsection{Multimodal Approaches}
To overcome the intrinsic limitations of single sensors, recent research has increasingly adopted multimodal fusion. Depth and Infrared (IR) sensors \citep{yin2022multimodal, tandon2024bodymap} provide surface geometry but are restricted to the visible line-of-sight. Notable efforts include Simultaneously-collected multimodal Lying Pose (SLP)-Fusion \citep{yin2022multimodal}, which integrates RGB, depth, pressure, and infrared signals to reconstruct the SMPL model under blankets. Similarly, BodyMAP \citep{tandon2024bodymap} jointly predicts body shape and pressure distribution from depth images, implicitly learning the correlation between surface geometry and contact forces. While fusing surface sensors improves robustness against noise and partial visibility, these methods are fundamentally limited to external observation. When a patient is mostly covered by surgical drapes or bedding, combined surface sensors (e.g., Depth + Pressure) still only capture the shape of the cover or the contact points. They cannot directly measure the occluded anatomy. Consequently, these systems must rely on estimating the missing geometry based on learned statistical correlations, which often fail to capture accurate patient-specific morphology. Furthermore, existing multimodal architectures typically rely on naive feature concatenation or early fusion, which treats all modalities with equal confidence. This approach fails to account for the varying reliability of sensors, particularly when one modality (e.g., depth) is occluded while another (e.g., the anatomical prior) remains reliable.

\subsection{Occlusion Handling and Anatomy-Guided Adaptation}
A critical barrier in clinical PSE is the severe occlusion caused by medical drapes. To address this without requiring large-scale labelled datasets of covered patients, two dominant strategies have emerged: synthetic data generation and domain adaptation.

Works such as BodyPressure \citep{clever2022bodypressure} and Patient MoCap \citep{achilles2016patient} utilise physics engines to simulate cloth interaction, training models on synthetic data. However, these models often suffer from a reality gap when deployed on real clinical data exhibiting complex drape deformations. To bridge this gap, Domain Adaptation (DA) techniques have been employed. \citet{chi2022multi} utilised adversarial learning, while Bigalke et al. \citet{bigalke2023anatomy} introduced anatomy-guided DA. The latter represents the current state-of-the-art for unlabelled adaptation, using \emph{generalised anatomical priors} such as limb symmetry and bone length ratios as weak supervision.

However, anatomy-guided DA exposes the central limitation of the current field: it assumes that patient conforms to a statistical norm. By enforcing symmetry or standard proportions, these methods risk imposing incorrect constraints on patients with pathological anomalies, scoliosis, or post-surgical asymmetry. This `one-size-fits-all' paradigm renders them unsuitable for high-precision surgical navigation. 

VIM-PSE addresses these compounding limitations by replacing generalised priors with a deterministic, patient-specific volumetric shape anchor, ensuring robust estimation for atypical anatomies without reliance on population-level statistical approximation. By fusing this occlusion-immune ground truth with depth-based pose data, our framework bridges the gap between surface-constrained sensing and the high-fidelity requirements of perioperative care.

\begin{figure}
\centering
\includegraphics[width=0.6\textwidth]{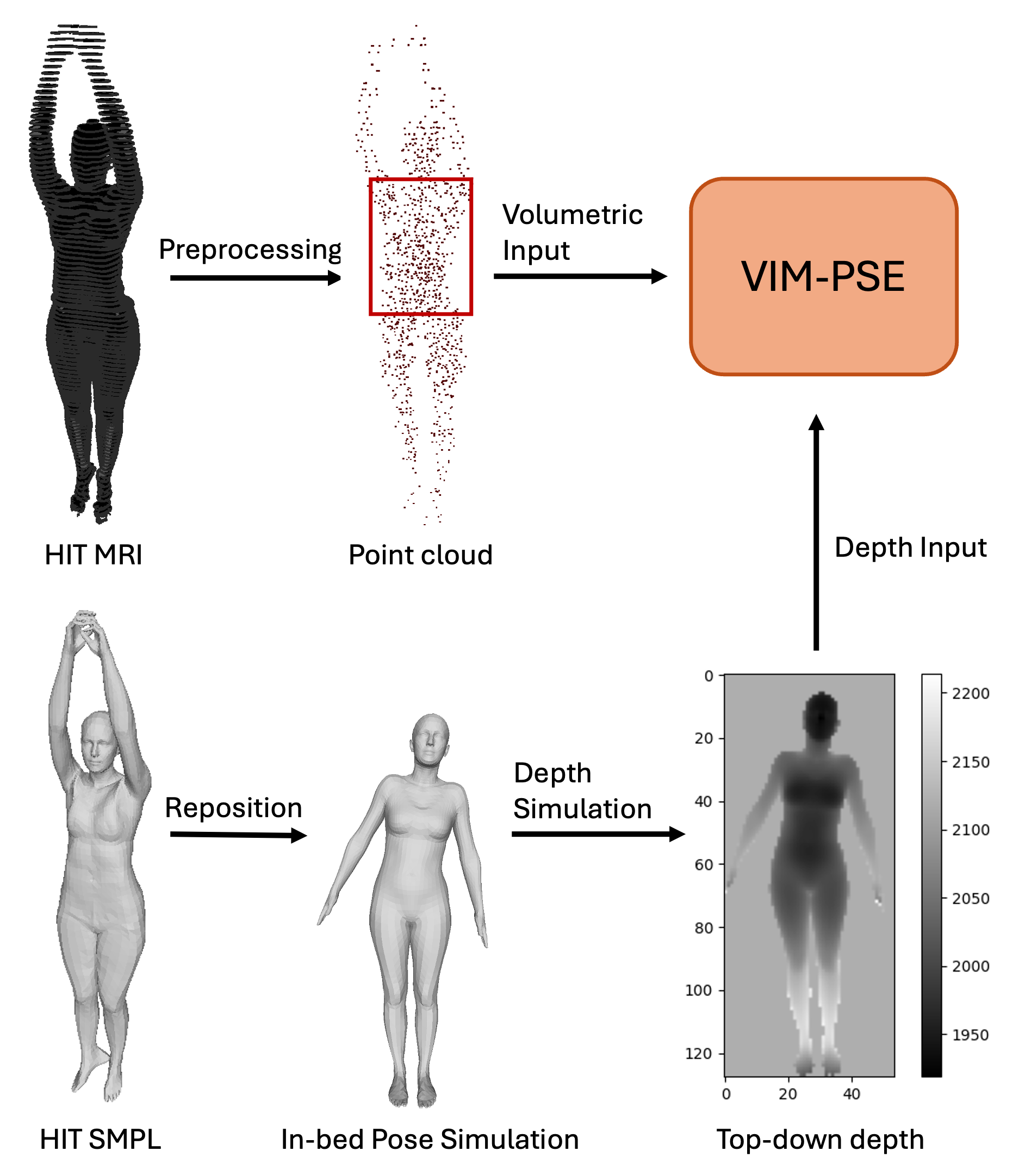}
\caption{Validation of VIM-PSE on the large-scale HIT dataset. A skin-surface point cloud is extracted from each subject's MRI scan to serve as a high-fidelity surrogate for the volumetric input required by our shape estimation module. To better match in-bed conditions, we applied in-bed poses from SLP before generating the simulated depth maps. This process creates an idealized pseudo ground truth, enabling direct comparison of predicted and known model parameters.} 
\label{fig3}
\end{figure}

\section{Methodology}
\label{sec:method}
\subsection{Problem Formulation and Notation}
\label{sec:problem_formulation}
The objective of VIM-PSE is to estimate the 3D pose and shape of a patient, represented by the SMPL parametric model parameters $\Theta$, given a multimodal input tuple. We define the training set as $\mathcal{D} = \{(X_{vol}^{(i)}, I_{depth}^{(i)}, \Theta_{gt}^{(i)})\}_{i=1}^{N}$.

The input consists of two distinct modalities:
\begin{enumerate}
    \item A patient-specific volumetric surface prior, represented as a point cloud $X_{vol} \in \mathbb{R}^{P \times 3}$ (where $P=5000$), which is extracted from the skin surface of a static volumetric medical scan (e.g., CT or MRI). This modality provides high-fidelity, occlusion-free shape information ($\beta$) but contains no dynamic pose information.
    \item A dynamic depth map $I_{depth} \in \mathbb{R}^{H \times W}$, which captures the top-down view of the patient in the hospital bed. This modality captures the current pose ($\theta$) but is subject to severe occlusion from bedding.
\end{enumerate}

The target output is the set of SMPL parameters $\Theta = \{\beta, \theta, R, \mathbf{t}\}$, where $\beta \in \mathbb{R}^{10}$ represents the principal component shape coefficients, $\theta \in \mathbb{R}^{72}$ denotes the axis-angle representations of the 24 body joints, and $\{R, \mathbf{t}\} \in \mathbb{R}^6$ represents the global orientation and translation. The SMPL function $\mathcal{M}(\beta, \theta)$ maps these parameters to a triangulated surface mesh $\mathcal{V} \in \mathbb{R}^{6890 \times 3}$ comprising $N_v = 6890$ vertices.

We formulate the inference problem as learning a mapping function $\mathcal{F}: (X_{vol}, I_{depth}) \rightarrow \hat{\Theta}$ that minimizes the deviation between the estimated model $\mathcal{M}(\hat{\Theta})$ and the ground truth geometry. Crucially, unlike unimodal approaches, we decompose the estimation into two optimized pathways: a shape-dominant pathway driven by $X_{vol}$ and a pose-dominant pathway driven by $I_{depth}$, fused via a cross-modal residual function. The overall learning objective is to find the optimal network parameters $\mathcal{W}^*$ that minimize the expected loss over the distribution:
\begin{equation}
\label{eq:objective}
\mathcal{W}^* = \arg\min_{\mathcal{W}} \mathbb{E}_{(X, I) \sim \mathcal{D}} [\mathcal{L}_{total}(\Theta_{gt}, \mathcal{F}(X_{vol}, I_{depth}; \mathcal{W}))]
\end{equation}
where $\mathcal{L}_{total}$ represents the combination of parametric regression, geometric alignment, and residual fusion losses detailed in the subsequent sections.

Our VIM-PSE model comprises two parallel modules: body shape estimation and pose estimation. The CRF module is then used to generate the final patient body mesh (see Fig.~\ref{fig2}). We use the SMPL model \citep{SMPL:2015} for 3D mesh reconstruction by leveraging its parametric representation, which decouples human body shape and pose into low-dimensional vectors. It includes pose $\theta$ $\in \mathbb{R}^{24\times3}$, shape $\beta$ $\in \mathbb{R}^{10}$, and global translation \textit{t} $\in \mathbb{R}^{3}$ parameters. In SMPL, $\beta$ is a vector of predefined principal component analysis shape coefficients that capture deviations from a standard T-shaped SMPL template. $\theta$ encodes relative 3D joint rotations in an axis-angle format to simulate pose changes.

\subsection{Body Shape Estimation Module}
A skin surface-segmented volumetric scan (e.g., CT or MRI) is first converted to a 3D surface mesh with marching cubes \citep{lorensen1998marching} and uniformly down-sampled to $N=5000$ points (see Module 1 in Fig.~\ref{fig2}).
The point cloud is encoded using PointNet++ \citep{qi2017pointnet++} as in \citet{zuo2021self}.  
For each point, the network outputs a local feature \(\mathbf{f}_n\) and a single global vector \(\mathbf{g}\) encoding the whole shape.  
Concatenating these, \([\mathbf{f}_n\|\mathbf{g}]\), two small multi-layer perceptron (MLP) heads predict a soft-assignment vector \(\boldsymbol{\pi}_{n}\) over the 6,890 template SMPL vertices and an outlier probability \(\mu_{n}\). These quantities appear directly in the Gaussian-mixture likelihood, as expressed in Eq.~\ref{eq:gmm}:  

\begin{equation}
\label{eq:gmm}
p(v_n)=(1-\mu_{n})\sum_{m=1}^{|M|}\pi_{mn}\,
\mathcal{N}\!\bigl(v_n \mid M_m(\Theta)\bigr)+\mu_{n}\tfrac{1}{N}
\end{equation}

where \(M_m(\Theta)\) is the \(m\)-th vertex of the SMPL mesh.
The parameter set \(\Theta=\{\beta,\theta,R,\mathbf{t}\}\) contains the
shape coefficients \(\beta\), joint poses \(\theta\), a global rotation \(R\),
and a global translation \(\mathbf{t}\). $\tfrac{1}{N}$ represents a uniform outlier probability over the volume of a bounding box.

A probabilistic-correspondence loop alternates (i) updating the posterior
match probabilities with the likelihood in (1), and (ii) minimizing the negative
log-likelihood with respect to \(\beta\) (holding \(\theta,R,\mathbf{t}\) fixed during
shape refinement).  Iteration converges to a shape that closely fits the torso
surface and yields the desired estimate of \(\beta\).

\subsection{Body Pose Estimation Module}
The pose estimation module uses a modified BodyMap network to exclusively process the top-view depth map of the patient to predict $\theta$ and \textit{t} \citep{tandon2024bodymap}. The original network takes both depth and pressure maps as inputs to output $\theta$, $\beta$, and \textit{t}. Our modified version exclusively uses a depth map and outputs only $\theta$ and \textit{t}, incorporating patient height information extracted from volumetric images to enhance accuracy. This modified BodyMap was adopted as the backbone architecture.

The input depth map is first normalized, denoised using a median filter, and resized to a shape of 128x54 in pixels. It is encoded using ResNet18 \citep{he2016deep} to extract latent features (denoted as $p_{depth}$) and processed through an MLP to regress $\theta$ and \textit{t}. This architecture is trained using a new loss function defined as Eq.~\ref{eq:loss}:

\begin{equation}
\label{eq:loss}
L \;=\; \frac{1}{2\sigma_{\text{smpl}}^{2}}\,L_{\text{SMPL}} \;+\;
          \frac{1}{2\sigma_{\text{v2v}}^{2}}\,L_{\text{v2v}} \;+\;
          \frac{1}{2\sigma_{\text{ht}}^{2}}\,L_{\text{height}}
\end{equation}
where $\sigma_{\text{smpl}},\sigma_{\text{v2v}},\sigma_{\text{ht}}$ are trainable scalars (initialized to 1.0) that implement homoscedastic-uncertainty weighting. This avoids arbitrary manual coefficients and balances losses with heterogeneous units.

\subsection{Cross-modal Residual Fusion Module}
To refine coherence between pose and shape predictions from volumetric data and depth, we introduce a lightweight CRF. The module follows prior work on residual fusion for multimodal learning~\citep{chen2020adafuse,deng2019rfbnet}, and operates directly in the SMPL parameter space. 

Given the initial shape prior $\beta_{vol}$ from CT and pose prior $\theta_{bm}$ from depth, we compute corrected estimates as:
\begin{equation}
\beta_{fused} = \beta_{vol} + \mathrm{MLP}_\beta([g_{vol}, p_{depth}])
\end{equation}
\begin{equation}
\theta_{fused} = \theta_{bm} + \mathrm{MLP}_\theta([p_{depth}, g_{vol}]),
\end{equation}
where $g_{vol}$ and $p_{depth}$ denote encoded features from the CT and depth pathways, and $[\cdot]$ is feature concatenation. Importantly, the modality-specific prior heads from CT and depth remain frozen, and the last layers of MLPs are zero-initialized, ensuring the fused predictions initially equal the priors, and residuals are learned only as small corrective offsets. The residual formulation ensures that each modality provides incremental refinements without overriding the priors. 

We supervise the fused predictions with ground truth SMPL parameters from the SLP dataset. The objective combines pose, shape, and joint-level accuracy:
\begin{equation}
L_{total} = \lambda_{pose} \cdot L_{pose} + \lambda_{shape} \cdot L_{shape} + \lambda_{joint} \cdot L_{joint},
\end{equation}
where $\lambda_{pose}, \lambda_{shape}, \lambda_{joint}$ are learnable weights optimized jointly with the network. The pose and shape losses are defined as:
\begin{equation}
L_{pose} = \frac{1}{K} \sum_{k=1}^{K} d_{geo}(R(\theta_{fused}^{(k)}), R(\theta_{gt}^{(k)}))
\end{equation}
\begin{equation}
L_{shape} = \|\beta_{fused} - \beta_{gt}\|_2^2,
\end{equation}
where $d_{geo}$ is the geodesic distance between rotation matrices $R(\cdot)$~\citep{zhou2019continuity}. For joint supervision, we focus on $N=8$ SMPL-defined torso joints, including neck, spine1-3, left and right shoulder, left and right collar:
\begin{equation}
L_{joint} = \frac{1}{N} \sum_{i=1}^{N} \| J_{pred}^{(i)} - J_{gt}^{(i)} \|_2,
\end{equation}
with $J_{pred}$ obtained via forward kinematics from the fused SMPL parameters and $J_{gt}$ provided by the SLP dataset.

The CRF module is trained end-to-end with Adam (learning rate $3 \times 10^{-4}$, weight decay $1 \times 10^{-5}$). A cosine annealing scheduler reduces the learning rate over 50 epochs. Batch size is 16. Each MLP uses two hidden layers with ReLU activation and dropout ($p=0.1$).

\section{Experimental Setup and Evaluation}
\subsection{Simulation Study}
To validate our approach on a large in silico cohort, we used the Human Implicit Tissues (HIT) dataset which contains $N=398$ adults (157 male, 241 female) scanned in the prone position (i.e., face down, lying front) using a 1.5 T MRI scanner (slice size $256\times192$, voxel size $\approx2\times2\times10$ mm\textsuperscript{3}) \citep{keller2024hit}. Each subject’s SMPL mesh was paired with their MRI-derived scan as shown in Fig.~\ref{fig3}. We selected this dataset because it provides paired volumetric medical imaging data and full-body SMPL registrations across a large and diverse population, making it well-suited for the synthetic generation of inputs for our task.  We randomly selected 300 subjects (150 male, 150 female) and generated two synthetic inputs required by the VIM-PSE pipeline: (i) a skin-surface point cloud derived from the reconstructed MRI mesh to serve as the volumetric input, and (ii) an orthographic top-view depth map from the paired ground truth SMPL mesh to simulate the bed-positioned patient. Patient height was calculated from the MRI-derived mesh bounding box and provided as the scalar input.

To ensure that the simulated measurements reflected realistic in-bed conditions, we re-posed all HIT subjects from their original scan posture into general in-bed poses. The target poses were taken from the publicly available SLP dataset, which captures natural supine sleeping configurations. We applied these SLP-derived pose parameters to the HIT SMPL registrations and regenerated the corresponding posed meshes before depth rendering. This step removes the scan-pose bias of HIT and aligns the simulated depth inputs with typical in-bed body configurations, as originally suggested during dataset construction to avoid overly idealised postures. Fig.~\ref{fig3} illustrates the re-posed HIT meshes using SLP-derived in-bed poses, which serve as the basis for generating our simulated depth inputs.

\begin{figure}
\centering
\includegraphics[width=0.6\textwidth]{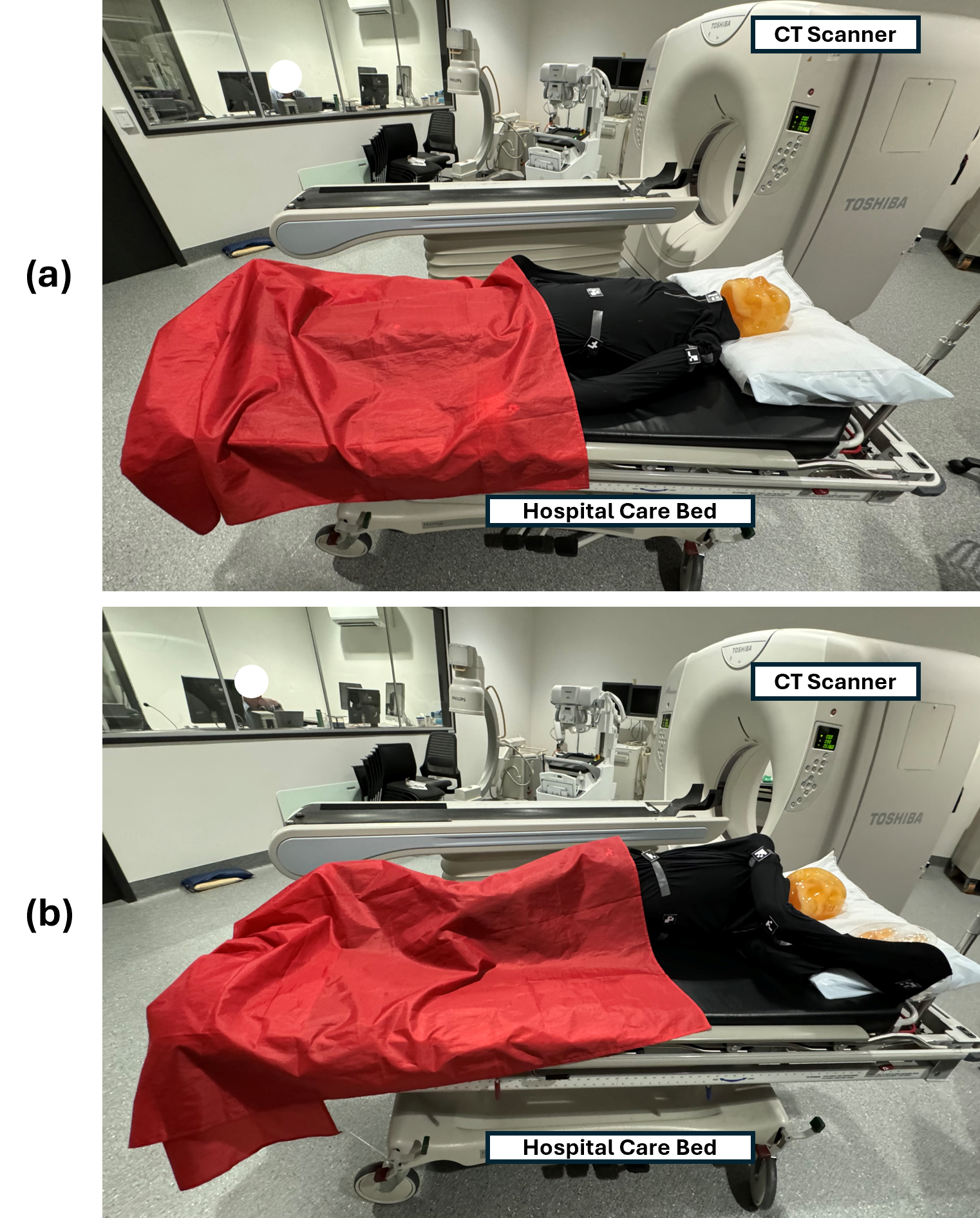}
\caption{The acquisition stage of the phantom dataset, with the phantom scanned on both the CT bed and the hospital care bed in two poses, supine (a) and lateral (b), both covered in the blanket.} 
\label{fig4}
\end{figure}

\subsubsection{MRI-Derived Skin Surface Extraction}
As VIM-PSE utilizes a generic volumetric surface prior, we reconstructed a comparable skin point cloud from MRI segmentation. Voxels with intensity values \( I(x,y,z) \geq T_{\mathrm{skin}} = 50 \) (12-bit scale) were first labelled as skin. To reduce noise, connected components smaller than 100 voxels were removed. From the resulting binary mask, an isosurface was extracted using the Marching Cubes algorithm to produce a triangulated skin mesh, which was then uniformly down-sampled to 5000 vertices, denoted as \( S_{\mathrm{skin}} \in \mathbb{R}^{5000 \times 3} \). To replicate the clinical workflow, \( S_{\mathrm{skin}} \) was cropped to the torso region with reference to paired SMPL landmarks, as stated in Section~\ref{sec:method}. All points outside these anatomical bounds were discarded, and the remaining torso surface was resampled to 1000 vertices, yielding the final set \( S_{\mathrm{torso}} \).

\subsubsection{Depth Map Simulation from SMPL Mesh}
For each HIT subject, we first adjusted the pose parameters in the ground truth SMPL with a known supine configuration (arms by side, legs extended) $\theta_{supine}$, to simulate a patient lying in bed as a pseudo ground truth (see Fig.~\ref{fig3}). We then placed a virtual orthographic camera above the SMPL mesh, projecting the SMPL vertices to produce a depth value per pixel. Any pixel with no projected vertex was assigned a fixed “bed” depth of 10 mm below the lowest body point, which was then normalized and cropped to a $240\times240$ array. The resulting depth image was used directly by the Pose-Estimation Module without further modification.

\subsection{Phantom and Volunteer Datasets}
The phantom dataset comprised four distinct configurations: two poses (supine, left-lateral), each with and without surgical drapes to mimic a realistic setting, using the PBU-60 whole-body phantom (Kyoto Kagaku Co., Kyoto, Japan). CT data of the phantom were acquired using a Toshiba Alexion CT scanner with a resolution of 512×512×1703 voxels (spacing: $0.761 \times 0.761 \times 0.8$\,mm$^3$). To generate the top‑view 2D depth map, the dressed phantom was placed on a hospital bed and scanned using a Microsoft HoloLens 2 (Microsoft Corporation, Redmond, WA, USA) to obtain a 3D surface (see Fig.~\ref{fig4} (a), (b)). The resulting mesh was projected to a 2D depth map via a virtual rendering camera located above the phantom using the same method as in the simulation study. Acquiring 3D scanning mesh enables accurate SMPL ground truth (GT) registration for each configuration. GT SMPL registration was obtained by aligning a SMPL model to the corresponding 3D HoloLens2 mesh using the algorithm in \citep{boja2020smpfitting}, followed by manual refinement and projection to 2D to align with depth maps. Quantification of the alignment was performed using the data loss (chamfer distance between the SMPL model and the 3D mesh), landmark loss (L2 distance between SMPL model landmarks and their correspondences in the mesh), prior shape loss (L2 norm of the SMPL shape parameters), and prior pose loss (GMM prior loss) to ensure a robust fit and to establish reliable GT. The final input to the shape estimation module was cropped to focus exclusively on the torso region, using the method described in Section~\ref{sec:method}. This region is of clinical significance because it is commonly affected by occlusion. For in-vivo validation, data were obtained from two healthy male volunteers with ethics approval from the Western Sydney Local Health District (2021/ETH00209): Volunteer 1 – height 172 cm, weight 77 kg, right lateral position; Volunteer 2 – 173 cm height, weight 70 kg, supine position (see Fig.~\ref{fig7}). Neither volunteer was covered with drapes. 

\begin{figure}
\centering
\includegraphics[width=0.4\textwidth]{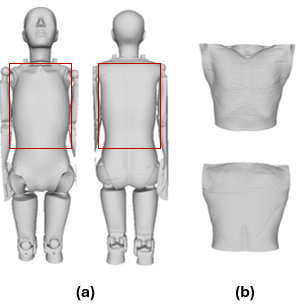}
\caption{The acquired CT-derived surface mesh for phantom (a) and the two volunteers (b), both shown as the front and back. The red rectangle on the phantom represents the cropped torso region used in this study.} 
\label{fig5}
\end{figure}

Thoracic CT data were acquired using a Vereos Digital PET/CT scanner. The resulting CT data for the phantom and the volunteers are shown in Fig.~\ref{fig5}. Top-view 2D depth maps and GT SMPL models were generated using the same approach as for the phantom dataset.

\subsection{Evaluation}
For the simulation study, we evaluated the difference between the SMPL models inferred from VIM-PSE and the corresponding pseudo ground truth SMPL models, as in Fig.~\ref{fig3}. 3D pose estimation performance was evaluated using the Mean Per-Joint Position Error (MPJPE) and Per-Vertex Error (PVE). MPJPE calculates the average Euclidean distance between the predicted 3D joint positions from the inferred SMPL model and those from the pseudo ground truth SMPL model. PVE calculates the average Euclidean distance between corresponding vertices of the predicted and GT SMPL models. To evaluate performance in 3D body shape estimation, anatomical accuracy was quantified by calculating the circumference of the chest, waist, and hips on the SMPL model, as proposed by \citet{choutas2022accurate}. These shape metrics were computed as the absolute error between the predicted and GT anatomical measurements. For the phantom and volunteer datasets we used the same metrics but compared the inferred SMPL from VIM-PSE directly with our ground truth SMPL.

To evaluate the impact of incorporating CT-derived $\beta$, we assessed V2V error between the inferred and ground truth SMPL model over the torso region. While the pose estimation network primarily takes depth input and estimates the full-body pose, we report only region-specific V2V errors for the chest area to better reflect the torso-specific robustness of VIM-PSE under occlusion and to avoid misleading aggregate metrics dominated by less relevant body parts (e.g., feet) in the 3D pose evaluation. 

\subsection{Ablation Studies}
We isolate the effect of the fusion policy between CT-anchored shape (Module~1) and depth-based pose (Module~2) by evaluating three regimes under the same protocols and metrics as ablation experiments. (i) \emph{Average($\beta,\theta$)}: we compute $\beta_{\text{avg}}=\tfrac{1}{2}(\beta_{\text{vol}}+\beta_{\text{depth}})$ and $\theta_{\text{avg}}=\tfrac{1}{2}(\theta_{\text{vol}}+\theta_{\text{depth}})$ while keeping $\mathbf{t}$ from Module~2, testing whether simple parameter averaging stabilizes reconstruction under occlusion. (ii) \emph{Concat}: we form the fused estimate by directly using the shape parameters from CT ($\beta=\beta_{\text{vol}}$ from Module~1) and the pose parameters from depth ($\theta=\theta_{\text{depth}}$, $\mathbf{t}$ from Module~2), without any residual corrections. (iii) \emph{CRF}: Cross-modal Residual Fusion with zero-initialized residual MLPs and learned confidence weights, so fused predictions initially match their modality priors and only small, data-driven corrections are learned.

\subsection{Baselines}
For the baseline comparison, we selected open-source state-of-the-art in-bed PSE methods that represent the current standards for geometric reconstruction and occlusion handling. BPBNet and BPWNet \citep{clever2022bodypressure} serve as the primary benchmarks for pressure-based estimation, BPBNet utilizes a purely data-driven residual architecture to regress SMPL parameters, while BPWNet integrates a physics-based white-box model to enforce contact constraints. We also include BodyMap-PointNet \citep{tandon2024bodymap}, which acts as a strong geometric baseline by jointly predicting body shape and surface pressure. Finally, we compare against AdaPose \citep{bigalke2023anatomy}, the current state-of-the-art for robust pose estimation under drapes, which employs anatomy-guided domain adaptation to align statistical priors with occluded observation data.

\begin{figure}
\centering
\includegraphics[width=0.6\textwidth]{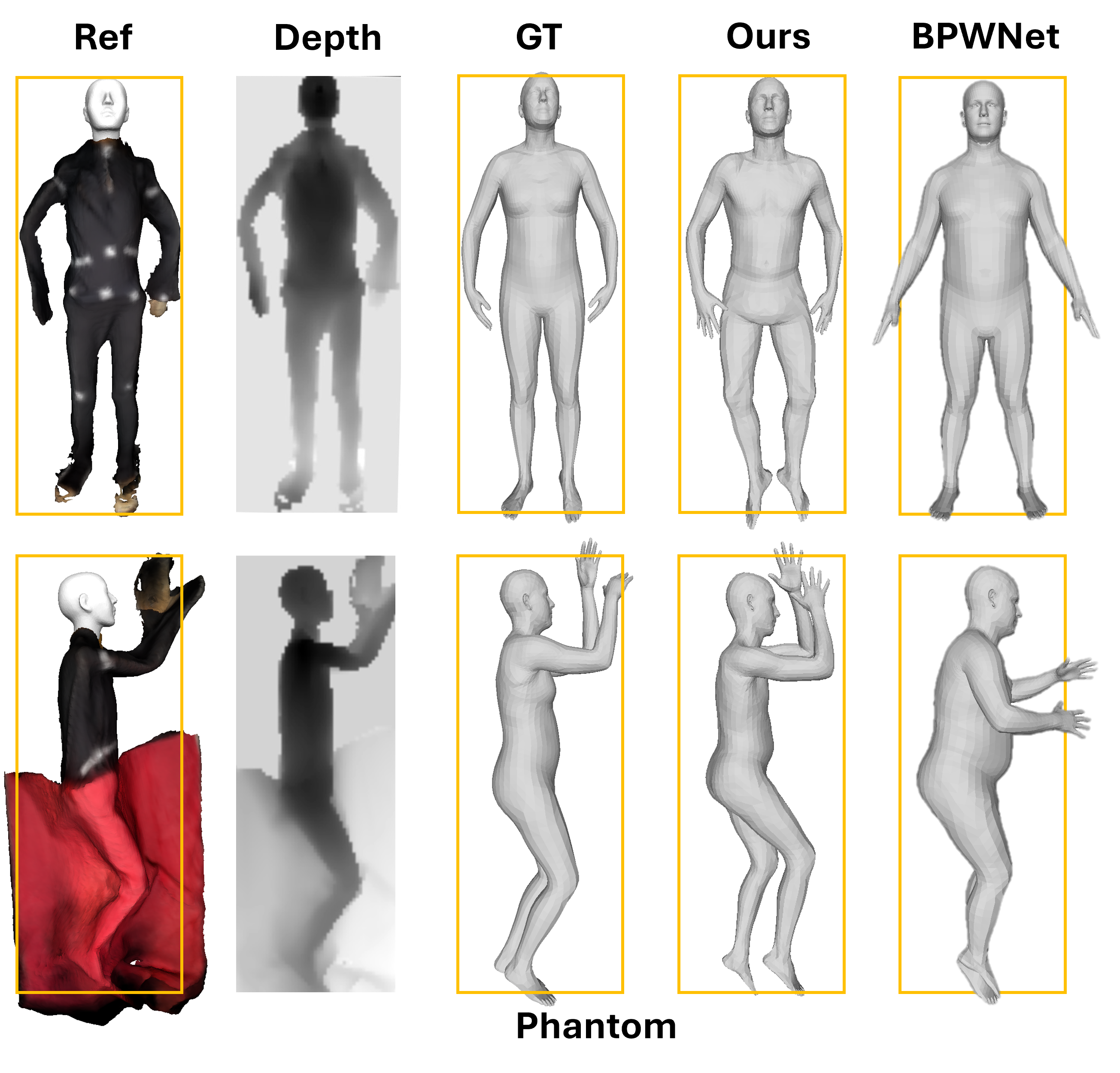}
\caption{Qualitative results on our phantom datasets in two positions. From left to right: reconstructed mesh, top-viewing depth map, GT SMPL model, VIM-PSE (ours) and BPWNet (baseline) results. The input CT is shown in Fig.~\ref{fig5}.} 
\label{fig6}
\end{figure}

\begin{figure}
\centering
\includegraphics[width=0.6\textwidth]{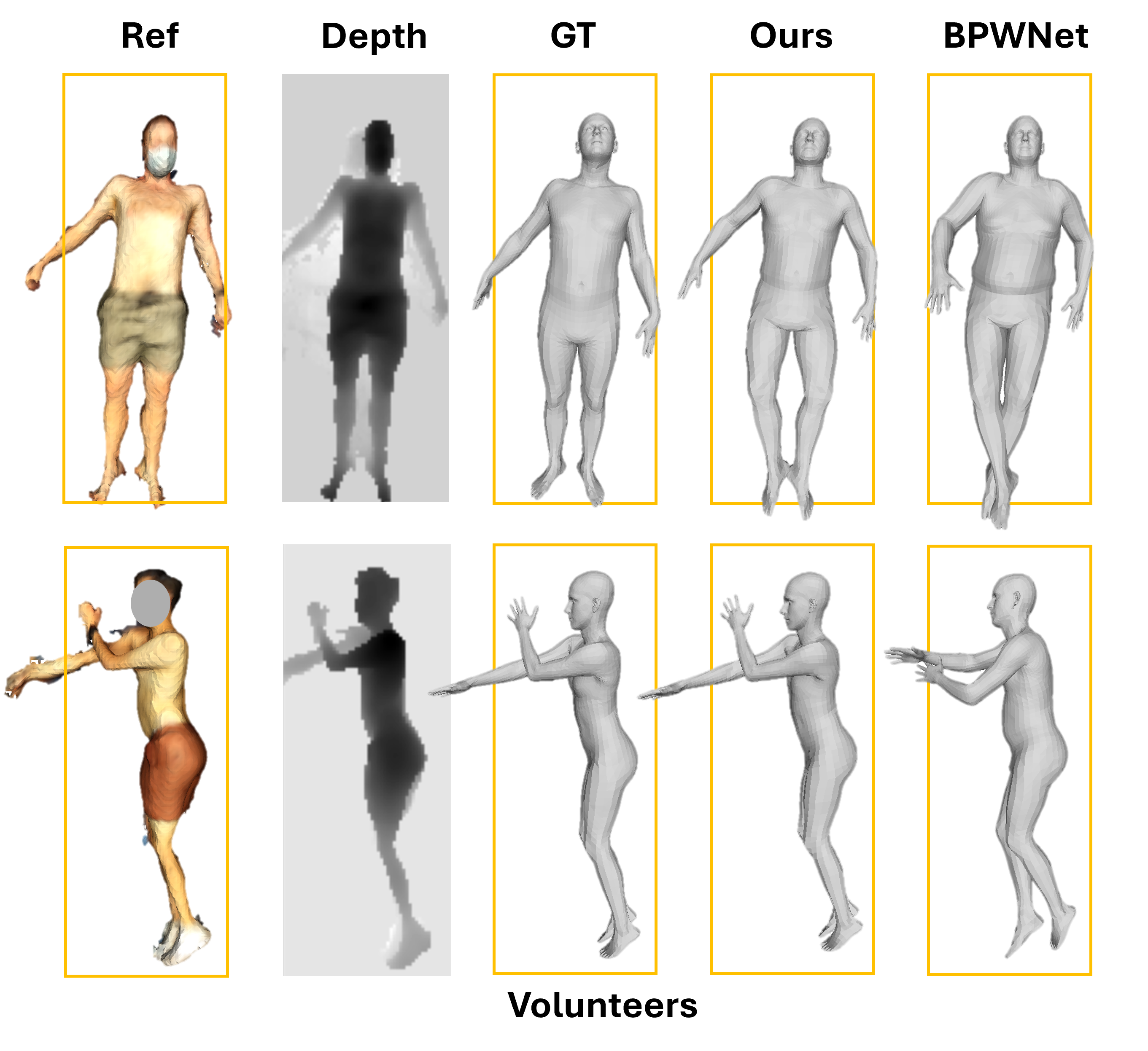}
\caption{Qualitative results on our proprietary volunteer datasets. From left to right: reconstructed mesh, top-viewing depth map, GT SMPL model, VIM-PSE (ours) and BPWNet (baseline) results. The two rows are from the two volunteers, and the input CT image is shown in Fig.~\ref{fig5}.} 
\label{fig7}
\end{figure}

\begin{table*}[t]
  \caption{Pose and shape errors on the simulated HIT dataset ($N=300$). The best results are in bold.}
  \label{tab:table1}
  \footnotesize
  \centering
  \begin{tabular}{lccccc}
    \toprule
    \multirow{2}{*}{Method} &
      \multicolumn{2}{c}{3D Pose Error (cm) ↓} &
      \multicolumn{3}{c}{3D Shape Error (cm) ↓} \\
    \cmidrule(lr){2-3} \cmidrule(lr){4-6}
    & MPJPE & PVE & Chest & Waist & Hip \\ \midrule
    BPBNet \citep{clever2022bodypressure}  
      & 9.31$\pm$0.28 & -- 
      & -- & -- & -- \\
    AdaPose \citep{bigalke2023anatomy}
      & 7.15$\pm$0.22 & -- 
      & -- & -- & -- \\
    BPWNet \citep{clever2022bodypressure}  
      & 8.12$\pm$0.25 & 9.88$\pm$0.22 
      & 10.11$\pm$0.30 & 18.75$\pm$0.33 & 13.02$\pm$0.38 \\
    BodyMap-PNet \citep{tandon2024bodymap} 
      & 6.92$\pm$0.20 & 7.80$\pm$0.19 
      & 5.76$\pm$0.25 & 4.79$\pm$0.22 & 5.90$\pm$0.23 \\
    VIM-PSE (Average)                
      & 4.11$\pm$0.12 & 4.92$\pm$0.11 
      & 1.90$\pm$0.32 & 2.06$\pm$0.15 & 1.83$\pm$0.12 \\
    VIM-PSE (Concat)                
      & 4.02$\pm$0.11 & 4.80$\pm$0.10 
      & 1.84$\pm$0.31 & \textbf{1.95$\pm$0.15} & 1.77$\pm$0.11 \\
    \textbf{VIM-PSE (CRF)}                
      & \textbf{3.98$\pm$0.10} & \textbf{4.75$\pm$0.09} 
      & \textbf{1.82$\pm$0.31} & 1.97$\pm$0.14 & \textbf{1.75$\pm$0.11} \\
    \bottomrule
  \end{tabular}
\end{table*}

\begin{table*}[t]
  \caption{Pose and shape errors on phantom and volunteer datasets ($N=6$) under draped and non-draped conditions. The best results for each condition are in bold. BPBNet and AdaPose produce 3D joint locations but not a surface mesh; therefore, only MPJPE is reported for these methods.}
  \label{tab:table2}
  \footnotesize
  \centering
  \renewcommand{\arraystretch}{1.05}
  \setlength{\tabcolsep}{3.8pt}
  \resizebox{\textwidth}{!}{
      \begin{tabular}{lccccccccccc}
        \toprule
        \multirow{3}{*}{Method} &
        \multicolumn{5}{c}{\textbf{Without Drape}} &
        \multicolumn{5}{c}{\textbf{With Drape}} \\
        \cmidrule(lr){2-6} \cmidrule(lr){7-11}
        & MPJPE ↓ & PVE ↓ & Chest ↓ & Waist ↓ & Hip ↓ 
        & MPJPE ↓ & PVE ↓ & Chest ↓ & Waist ↓ & Hip ↓ \\ 
        \midrule
        BPBNet \citep{clever2022bodypressure}  
          & 8.93$\pm$0.30 & -- & -- & -- & -- 
          & 10.82$\pm$0.41 & -- & -- & -- & -- \\
        AdaPose \citep{bigalke2023anatomy}
          & 7.92$\pm$0.24 & -- & -- & -- & -- 
          & 8.21$\pm$0.27 & -- & -- & -- & -- \\
        BPWNet \citep{clever2022bodypressure}  
          & 7.65$\pm$0.25 & 9.12$\pm$0.21 & 9.82$\pm$0.32 & 19.59$\pm$0.34 & 12.81$\pm$0.42 
          & 9.48$\pm$0.33 & 11.27$\pm$0.28 & 14.63$\pm$0.39 & 23.42$\pm$0.43 & 15.10$\pm$0.49 \\
        
        BodyMap-PNet \citep{tandon2024bodymap} 
          & 6.37$\pm$0.18 & 7.30$\pm$0.18 & 5.38$\pm$0.24 & 4.36$\pm$0.21 & 5.57$\pm$0.21 
          & 7.85$\pm$0.22 & 8.91$\pm$0.20 & 7.42$\pm$0.28 & 6.03$\pm$0.25 & 6.92$\pm$0.23 \\
        VIM-PSE (Average)              
          & 4.96$\pm$0.14 & 5.92$\pm$0.12 & 2.55$\pm$0.44 & 2.79$\pm$0.19 & 2.38$\pm$0.13 
          & 5.21$\pm$0.15 & 6.28$\pm$0.13 & 2.71$\pm$0.45 & 2.93$\pm$0.20 & 2.52$\pm$0.13 \\
        VIM-PSE (Concat)              
          & 4.86$\pm$0.13 & 5.77$\pm$0.11 & 2.47$\pm$0.43 & 2.69$\pm$0.18 & 2.31$\pm$0.12 
          & 5.08$\pm$0.14 & 6.09$\pm$0.12 & 2.62$\pm$0.44 & 2.83$\pm$0.19 & 2.43$\pm$0.12 \\
        \textbf{VIM-PSE (CRF)}                
          & \textbf{4.82$\pm$0.13} & \textbf{5.73$\pm$0.11} & \textbf{2.46$\pm$0.43} & \textbf{2.67$\pm$0.18} & \textbf{2.30$\pm$0.12} 
          & \textbf{4.95$\pm$0.13} & \textbf{5.86$\pm$0.11} & \textbf{2.53$\pm$0.43} & \textbf{2.74$\pm$0.18} & \textbf{2.35$\pm$0.12} \\
        \bottomrule
      \end{tabular}
      }
\end{table*}

\section{Results and Discussion}
\subsection{Simulation Study}
The HIT dataset enables direct comparison between the predicted and known SMPL pose and shape parameters. As shown in Table~\ref{tab:table1}, VIM-PSE with CRF achieved an MPJPE of 3.98$\pm$0.10 cm and a PVE of 4.75$\pm$0.09 cm. In comparison to the baselines, VIM-PSE significantly outperformed the pressure-based methods BPBNet (9.31 cm) and BPWNet (8.12 cm), the domain-adaptation method AdaPose (7.15 cm), and the second-best performing method, BodyMap-PNet (6.92 cm), improving upon the latter by 42.5\% (MPJPE) and 39.1\% (PVE). For shape estimation, VIM-PSE with CRF further reduced errors to 1.82$\pm$0.31 cm (chest), 1.97$\pm$0.14 cm (waist), and 1.75$\pm$0.11 cm (hip).

This performance advantage stems from the fundamental limitations of the baseline approaches. Methods like BodyMap-PNet rely on learning implicit correlations between surface depth and body shape; in the simulation, these correlations are weakened when bedding occludes the true surface. Similarly, while AdaPose improves over standard baselines by aligning feature distributions, it still lags significantly behind VIM-PSE because statistical domain adaptation cannot compensate for the lack of patient-specific geometric data. By explicitly incorporating a volumetric prior, VIM-PSE resolves the ambiguity that purely data-driven methods fail to capture.

\subsection{Phantom and Volunteer Study}
\subsubsection{Pose and Body Shape Estimation}
Table~\ref{tab:table2} summarizes results across phantom and volunteer configurations. Relative to the second-best baseline BodyMap-PNet, the VIM-PSE (CRF) variant reduced MPJPE from 6.37 $\pm$ 0.18\,cm to 4.82 $\pm$ 0.13\,cm (24.3\% improvement) and PVE from 7.30 $\pm$ 0.18\,cm to 5.73 $\pm$ 0.11\,cm (21.5\% improvement). Shape errors also dropped substantially, with chest error reducing by 54.3\% (5.38 to 2.46\,cm). These gains confirm the benefit of volumetric-anchored torso geometry under realistic occlusion and noise.

Regarding the ablation strategies (Table~\ref{tab:table2}), the CRF variant consistently achieved the lowest errors across all metrics, followed closely by Concat, with Average performing the worst. The performance gap between Concat and CRF remains intentionally small (<1\%). This marginal difference suggests that the primary robustness of VIM-PSE is driven by the high-fidelity volumetric priors themselves, which stabilize the solution significantly even with simple concatenation. The CRF module provides fine-grained residual tuning rather than gross structural correction, validating the design goal of a lightweight fusion coupler.

\subsubsection{With and Without Drape}
As shown in Table~\ref{tab:table2}, all methods experience performance degradation under the draped condition due to partial occlusions. However, VIM-PSE exhibits markedly higher robustness compared to prior approaches. For instance, BPWNet and BodyMap-PNet show noticeable increases in pose and shape errors when transitioning to the draped setting; notably, the waist error for BPWNet increased by 3.83\,cm. This degradation occurs because depth-based methods inevitably interpret the raised surface of the drape as a larger body volume, leading to overestimated shape parameters ("bloating") and misaligned joints.

In contrast, VIM-PSE demonstrates only a marginal increase in error under occlusion (MPJPE: 4.82$\rightarrow$4.95\,cm), maintaining sub-3\,cm accuracy across all anatomical regions. The underlying reason for this robustness lies in the integration of the volumetric structural prior, which effectively `locks' the torso shape and preserves anatomical relationships even when the superficial geometry is distorted by drapes. The CRF-based refinement further contributes by enforcing local spatial coherence, ensuring that the inferred mesh does not fragment across occluded boundaries.

\begin{figure}
\centering
\includegraphics[width=0.6\textwidth]{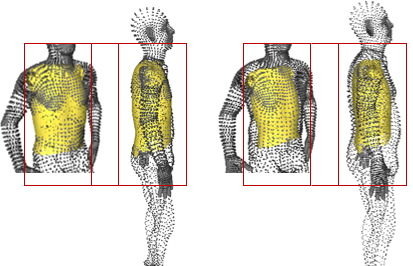}
\caption{The torso-specific case study compares our SMPL output (left) against the BodyMap-PointNet’s SMPL output (right). The yellow mesh represents the raw phantom CT, which aligns more closely in our SMPL output, better representing the body shape.} 
\label{fig8}
\end{figure}

\subsubsection{Qualitative Results}
Figs.~\ref{fig6} and ~\ref{fig7} illustrate the qualitative performance of VIM-PSE on the phantom and volunteer datasets compared with the ground truth SMPL and BPWNet. In the phantom example with surgical drapes (second row of Fig.~\ref{fig6}), VIM-PSE successfully infers the occluded body structure, whereas BPWNet struggles to predict the body shape, often aligning to the drape surface rather than the underlying anatomy. Similar accuracy was observed in the volunteer cases (Fig.~\ref{fig7}), where VIM-PSE generalizes well to human anatomy despite soft tissue variability.

However, two limitations are evident from the visual comparisons. First, discrepancies are visible in the lower limbs, particularly in leg contours, likely due to the lack of volumetric coverage below the pelvis in our datasets. As a result, while the torso is reliably reconstructed, the legs remain dependent on depth input alone. Second, subtle artifacts appear in arm alignment when arms are outstretched (e.g., second row of Fig.~\ref{fig7}), potentially due to the limited top-down field-of-view. These observations suggest that while volumetric geometry enhances torso estimation, distal limb accuracy requires expanded scan coverage or additional camera views.

\begin{table}[t]
  \begin{threeparttable}
    \caption{Torso-specific V2V error on phantom and volunteer poses. The best results are in bold.}
    \label{tab:table3}
    \footnotesize
    \centering
    \begin{tabular*}{\linewidth}{@{\extracolsep{\fill}}lcccc}
      \toprule
      \multirow{2}{*}{Configuration} & \multicolumn{4}{c}{V2V (cm) ↓} \\ \cmidrule(l){2-5}
         & BPBNet & BPWNet & BodyMap & VIM-PSE \\
      \midrule
      P\tnote{a}-Back           & 0.36$\pm$0.24 & 0.37$\pm$0.22 & 0.29$\pm$0.22 & \textbf{0.22$\pm$0.13} \\
      P\tnote{a}-Back, draped   & 0.43$\pm$0.26 & 0.44$\pm$0.25 & 0.31$\pm$0.23 & \textbf{0.24$\pm$0.15} \\
      P\tnote{a}-L-side\tnote{b} & 0.51$\pm$0.31 & 0.50$\pm$0.28 & 0.35$\pm$0.24 & \textbf{0.26$\pm$0.12} \\
      P\tnote{a}-L-side, draped & 0.48$\pm$0.28 & 0.49$\pm$0.26 & 0.37$\pm$0.25 & \textbf{0.30$\pm$0.18} \\
      \midrule
      V\tnote{c}-Back           & 0.35$\pm$0.21 & 0.35$\pm$0.21 & 0.33$\pm$0.26 & \textbf{0.25$\pm$0.15} \\
      V\tnote{c}-R-side\tnote{d} & 0.47$\pm$0.28 & 0.46$\pm$0.25 & 0.39$\pm$0.27 & \textbf{0.27$\pm$0.19} \\
      \bottomrule
    \end{tabular*}
    
        \begin{tablenotes}
      \item \textsuperscript{a}P: Phantom dataset. \quad \textsuperscript{b}L-side: Left-lateral orientation. \quad \textsuperscript{c}V: Volunteer dataset. \quad \textsuperscript{d}R-side: Right-lateral orientation.
    \end{tablenotes}
    
  \end{threeparttable}
\end{table}

\subsubsection{Torso-specific analysis}
Table~\ref{tab:table3} decomposes V2V errors for different configurations. Overall, body orientation and occlusion markedly affect accuracy. On the phantom, the lowest torso V2V was observed in the supine position (0.22$\pm$0.13\,cm with VIM-PSE), consistent with high torso visibility and stable bed contact. Lateral poses exhibited slightly higher V2V for all methods, reflecting greater difficulty when estimating body contour from oblique angles. Across all orientations, drapes increased error for baselines significantly, whereas VIM-PSE maintained stability, highlighting its capacity for robust reconstruction in realistic settings.

\subsection{Limitations and Future Work}
Our primary limitation is the small sample size (N=6) of our real-world phantom and volunteer datasets. To mitigate this lack of anatomical variability, we validated our core methodology on the large-scale simulated HIT dataset (N=300), which covers a diverse range of body shapes and sizes. However, this simulation operates under idealized conditions with perfectly registered data. Therefore, future work will involve a large-scale, multi-centre clinical validation to assess the robustness of VIM-PSE across diverse patient demographics and real-world sensor noise.

A second limitation is the dependence of our method on pre-acquired volumetric scans. Although this provides high-fidelity priors that stabilize shape estimation, it limits the method’s use to contexts involving preoperative imaging. Future work will investigate estimating or synthesizing volumetric-like representations directly from non-radiological modalities, such as SMPL-based skeletal models, to emulate the structural constraints offered by VIM-PSE. This approach could enable the framework to generalize beyond contexts requiring full volumetric imaging, providing an anatomically informed prior compatible with low-cost acquisition systems.

We also identify potential clinical applications which are beyond the scope of this paper but will be considered for future work. First, AR systems in surgical navigation require strict spatial registration, often demanding entry-point tolerances of 1--3\,mm for procedures like pedicle screw placement~\citep{ElmiTerander2016_AR_SPINE,Spirig2021_AR_Pedicle}. Our current evaluation (torso V2V error of 0.26\,cm) suggests that VIM-PSE approaches this high-precision requirement, warranting further investigation into its intraoperative viability. Second, accurate surface geometry is central to proactive pressure-injury prevention, where clinical guidance often targets limiting unintended displacement to $\approx$2\,cm~\citep{IslandHealth2018_Positioning,NPIAP2019_Guideline}. Our preliminary circumference errors ($\approx$2.48\,cm) indicate that VIM-PSE could provide actionable cues for detecting deviations relevant to pressure-care planning. Future studies will focus on deploying our framework in these specific clinical downstream tasks to validate its operational utility.

\section{Conclusion}
In this study, we proposed VIM-PSE, a novel multimodal network that integrates volumetric medical data and depth maps to enhance the accuracy of in-bed patient PSE. Our network leverages the imaging data routinely acquired in perioperative settings to optimize shape estimation. We further introduced a lightweight Cross-modal Confidence-Guided Module that enables consistent performance refinements through confidence-weighted feature corrections. Pose and shape evaluation on simulations and real-world (phantom and volunteer) datasets demonstrated that VIM-PSE consistently surpassed state-of-the-art counterparts, particularly under challenging occlusion scenarios. These improvements yield precise 3D patient mesh models that have the potential to enhance clinical workflows, including patient positioning, AR-guided surgical navigation, and postoperative monitoring.

\section{Acknowledgment}
The authors thank Assoc. Prof. Peter Kench and Mr Peter O’Reilly from the School of Health Sciences, The University of Sydney for their help and support during the acquisition of the whole-body phantom and volunteer dataset.

\section*{Declaration of generative AI and AI-assisted technologies in the manuscript preparation process}

During the preparation of this work the author(s) used Gemini 3.0 Pro in order to improve the readability and language of the manuscript. After using this tool/service, the author(s) reviewed and edited the content as needed and take(s) full responsibility for the content of the published article.
\bibliographystyle{elsarticle-num}
\bibliography{references}

\end{document}